\documentclass{article}

\usepackage[final]{nips_2016}

\usepackage[utf8]{inputenc} 
\usepackage[T1]{fontenc}    
\usepackage{hyperref}       
\usepackage{url}            
\usepackage{booktabs}       
\usepackage{amsfonts}       
\usepackage{microtype}      
\usepackage[pdftex]{graphicx}
\usepackage{amsmath}
\usepackage{subfig}
\usepackage[font={small}]{caption}

\title{LinXGBoost: Extension of XGBoost to\\Generalized Local Linear Models}

\author{
  Laurent de Vito \\ 
  \texttt{devitolaurent@gmail.com} \\
}

\begin{document}

\maketitle

\begin{abstract}
XGBoost is often presented as the algorithm that wins every ML competition.
Surprisingly, this is true even though predictions are piecewise constant.
This might be justified in high dimensional
input spaces, but when the number of features is low, a piecewise linear model
is likely to perform better. XGBoost was extended into LinXGBoost that stores
at each leaf a linear model. This extension, equivalent to piecewise
regularized least-squares, is particularly attractive
for regression of functions that exhibits jumps or discontinuities.
Those functions are notoriously hard to regress. Our extension
is compared to the
vanilla XGBoost and Random Forest
in experiments on both synthetic
and real-world data sets.
\end{abstract}

\section{Introduction}

Most competitors in ML jump straight to XGBoost,
\citet{DBLP:journals/corr/ChenG16}, an implementation of
the gradient boosting algorithm, because of its speed and accurate predictive
power. Part of XGBoost amazing speed must be ascribed to the fact that
predictions are piecewise constant. From a modeling perspective, this might certainly
be the right thing to do in high-dimensional input spaces, but if the number of features
is low, a piecewise linear model is likely to yield a better predictive performance.

\begin{figure}[h]
  \centering
  \includegraphics[width=1.0\linewidth]{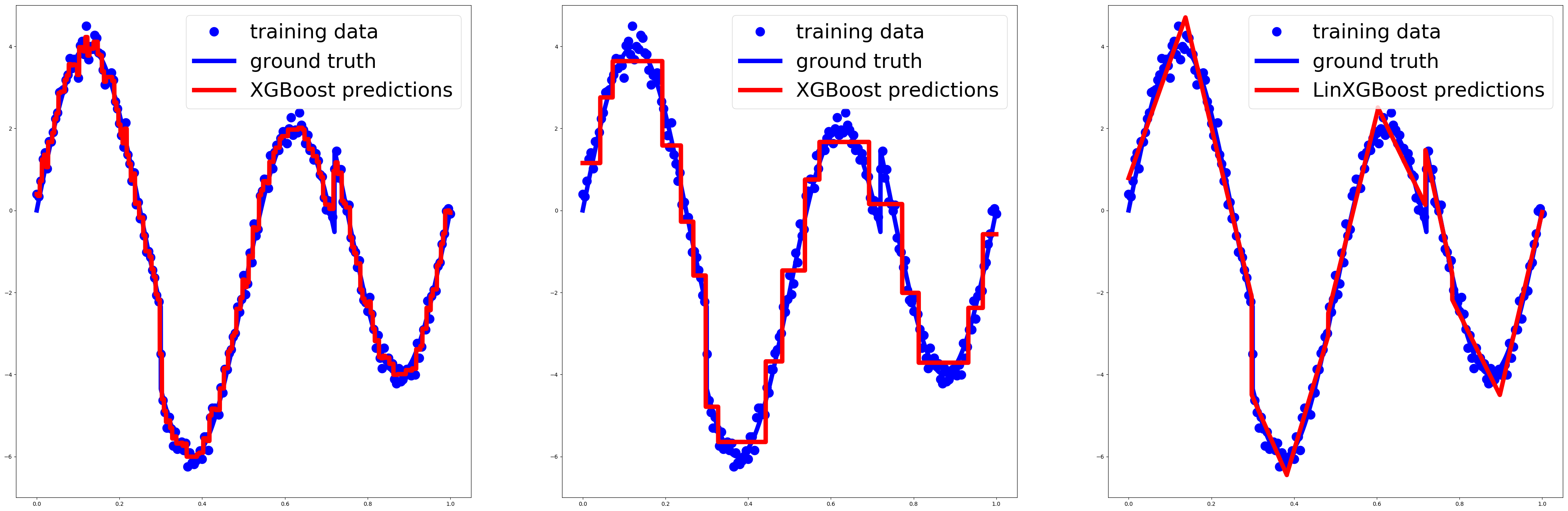}
  \caption{Regression of the \emph{HeavySine} function based on 201 noisy
  samples (the noise is Gaussian with zero mean and variance $\sigma^2 = 0.05$).
  Left: XGBoost with default setting except that the learning rate is reduced
  to 0.1 and the number of trees boosted to 50.
  Middle: XGBoost using a single tree.
  Right: LinXGBoost using also a single tree.
  In the middle and right plots, both XGBoost and LinXGBoost share the same
  tree maximum depth, 30, and regularization term on the number
  of leaves $\gamma=3$ and L2 regularization term on weights $\alpha=0$.}
  \vspace{-25pt}
  \label{fig: many trees vs. single tree}
\end{figure}

This is best seen on a one-dimensional function. Consider the synthetic
\emph{HeavySine} function, \citet{donoho1995adapting},
a sinusoid of period 1 with two jumps, at $t_1=.3$ and $t_2=.72$:
\begin{equation}
f(t)=4 \sin(4 \pi t) - \text{sign}(t-0.3)-\text{sign}(0.72-t)
\end{equation}
Given enough trees, XGBoost can adequately fit the noisy data set,
Figure \ref{fig: many trees vs. single tree}, left.
If we constrain XGBoost to a single tree and further regularize the model
to prevent over-fitting, the piecewise constant nature of the
predictions is clearly revealed, Figure \ref{fig: many trees vs. single tree},
middle. A single tree with a linear model at the leaves produces visually
far better results, Figure \ref{fig: many trees vs. single tree}, right.
To get better results in terms of the NMSE performance metric,
more trees are needed though.
By adding quadratic terms, we can even get superior results.

A piecewise (constant or linear) model is particularly suited
for the regression of functions
that exhibit jumps or discontinuities.
Finding jumps in otherwise smooth functions is a notoriously hard challenge.
It usually involves the derivation of criteria for choosing the number
and placement of the jumps, \citet{lee2002automatic}.
Even Bayesian models treat exclusively jumps in one-dimensional signals,
i.e. time series, \citet{adams2007bayesian}.

A piecewise linear model is appropriate for functions whose smoothness
is input-dependent. There are well established techniques
to regress functions whose smoothness does
not vary (e.g. the Nadaraya-Watson kernel estimator or Gaussian processes,
both with a fixed bandwidth). While convenient, the assumption that the
smoothness is input-independent is rarely realistic. Extensions to model
functions with varying length-scales are not trivial (e.g. \emph{adaptive}
Nadaraya-Watson kernel estimator or \emph{non-stationary} Gaussian processes).
Nevertheless, this is essential in many fundamental problems.
Modeling terrain surfaces, for instance, given sets of noisy elevation
measurements is even more challenging since it requires the ability
to balance smoothing against the preservation of discontinuities
(see e.g. \citet{plagemann2008nonstationary}).

At the core of gradient boosting are regression trees. A regression tree
decomposes the input space
into subdomains. This is reminiscent of the \emph{domain decomposition} approach
to regress non-stationary functions, \citet{park2011domain}.
In each input subdomain, the function can be approximately regarded
as stationary. Therefore, local regression is applied in each subdomain with
a fixed length-scale or bandwidth. But this results in discontinuities in
prediction on boundaries of the subdomains, \citet{JMLR:v17:15-327}.
To mitigate this drawback,
the assignment of data points to models can also be soft:
The assignments are treated as unobserved random variables,
\citet{rasmussen2002infinite}.
Because of the uncertainty in the assignments,
discontinuities in predictions are smoothed out.

Domain decomposition is however rarely done in a principled way: In
\citet{kim2005analyzing}, the uncertainty in the number of
disjoint regions, their shapes and the model within regions was dealt with in
a fully Bayesian fashion. However, this feat has a price: It is fairly
involved and slow (because of Reversible jump MCMC) and limited to small
feature space dimensions (because of the Voronoi tessellation).

Finding the right size of the local region for linearization is a problem faced
in Locally Weighted Learning, \citet{englert2012locally}.
\citet{meier2014local} developed a Bayesian
localized regression algorithm. The local models were added incrementally
but not in a fully consistent manner.

\begin{figure}[h]
  \centering
  \includegraphics[width=1.0\linewidth]{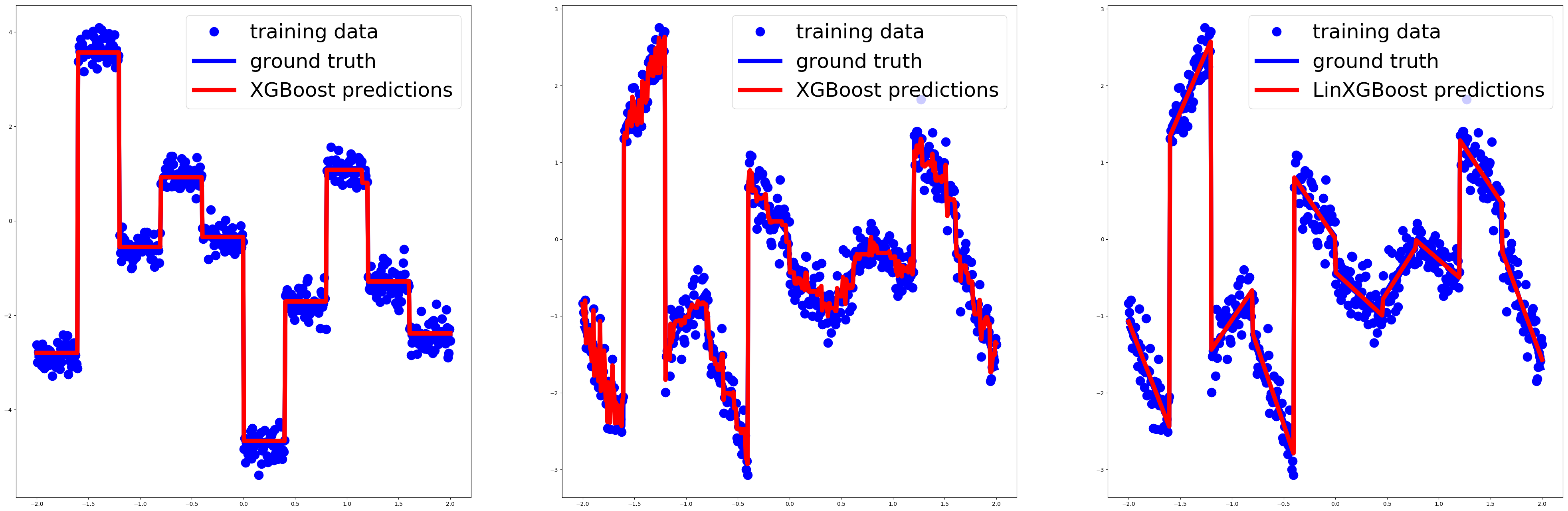}
  \caption{Left: Regression of a noisy data set from a piecewise constant function.
  XGBoost can automatically infer the right number of partitions using a single
  tree.
  Middle: Regression of a noisy data set from a piecewise linear function.
  Because XGBoost predictions are piecewise constant,
  we are forced to boost the number of trees (and so reduce the learning rate)
  to fit the data. The tree was not regularized: A strong regularization would
  lead to piecewise constant predictions.
  Right: LinXGBoost using a single tree provides a much better match.
  Notice that the tree was regularized.}
  \label{fig: pc vs. pl}
\end{figure}

Though XGBoost is not Bayesian, the trees are grown and the scores at the
leaves are chosen in a principled way: To minimize an objective function.
Consequently, XGBoost can automatically captured jumps and discontinuities,
Figure \ref{fig: pc vs. pl}, left.
Through the extension to local \emph{linear} models, we make it able to
additionally better model smooth functions with varying length-scales
using fewer trees, Figure \ref{fig: pc vs. pl}, middle and right.

It is not possible to natively handle categorical features;
as in XGBoost, they must be encoded into numerical vectors
using e.g. one-hot encoding.

Notice that plugging-in higher-order models at the tree leaves
was advocated by \citet{torgo1997functional}
to produce \emph{local regression trees}. He acknowledged that
this strategy brings significant gains in terms of prediction accuracy
at the cost of an increase of computation time.

\section{The vanilla XGBoost}
\label{bosting}

We follow and borrow material from the clear-cut presentation in \citet{chen2014introduction}.
We refer to this presentation for further details.

XGBoost implements a Boosting algorithm.
Boosting algorithms belongs to \emph{ensemble machine learning methods}.
Specifically, they iteratively add predictors that focus on improving
the current model, and this is achieved by modifying the learning problem
(the objective in XGBoost) between iterations, see e.g.
\citet{elith2008working}.
Hence, boosting algorithms greedily approximate a function.

The predictors XGBoost builds are \emph{regression trees}.
A regression tree has
decision rules and scores at the leaves.
It is a function since it maps features (the attributes) to values (the scores).
The prediction at $\boldsymbol{x}_*$ is given by
\begin{equation}
\hat{y} = \sum_{k=1}^K{ f_k(\boldsymbol{x}_*) }
\end{equation}
assuming we have $K$ trees.
$f_k \in \mathcal{F}$ where $\mathcal{F}$ is the space of functions
containing all regression trees.
The set of parameters is thus $\boldsymbol{\theta} = \{ f_1, \cdots, f_K \}$.
Instead of learning weights as in linear regression or logistic regression,
we are learning functions.

In XGBoost, the learning of functions is done by defining
an objective to minimize:
\begin{equation}
Obj = \sum_{i=1}^n{ \ell(y_i, \hat{y}_i) } + \sum_{k=1}^K{ \Omega_k(f_k) }
\end{equation}
where $\hat{y}_i$ is the prediction and $y_i$ the observation
at $\boldsymbol{x}_i$ for $i=1,\cdots,n$.
$\ell(\cdot,\cdot)$ designates a loss function.
The first term is the training loss,
the second penalizes the complexity of trees.

In XGBoost, the objective at step \(t\) is defined as
\begin{eqnarray}
Obj^{(t)} &=& \sum_{i=1}^n{ \ell(y_i, \hat{y}_i^{(t)}) } + \sum_{k=1}^t{ \Omega_k(f_k) } \\
              &=& \sum_{i=1}^n{ \ell(y_i, \hat{y}_i^{(t-1)}+f_t(\boldsymbol{x}_i)) } + \sum_{k=1}^t{ \Omega_k(f_k) }
\end{eqnarray}
$f_t(\boldsymbol{x}_i)$ can be thought of as a perturbation (the residual),
and so the loss is, to second-order accuracy:
\begin{equation}
\ell(y_i, \hat{y}_i^{(t-1)}+f_t(\boldsymbol{x}_i))
\approx
\ell(y_i, \hat{y}_i^{(t-1)})
+ g_i f_t(\boldsymbol{x}_i)
+ \frac{1}{2} h_i  f_t(\boldsymbol{x}_i)^2
\end{equation}
where $g_i$ (resp. $h_i$) is the first (resp. second) derivative of
the loss w.r.t. its second argument evaluated at $(y_i, \hat{y}_i^{(t-1)})$.
For the square loss, $\ell(y_i, \hat{y})=(y_i- \hat{y})^2$,
we have $g_i=2 (\hat{y}_i^{(t-1)}-y_i)$ and $h_i=2$.
In that particular case, the approximation is exact.

Removing the constant terms, we get
\begin{equation}
\label{obj at round t}
Obj^{(t)}
=
\sum_{i=1}^n{
                    [ g_i f_t(\boldsymbol{x}_i)) + \frac{1}{2} h_i  f_t(\boldsymbol{x}_i)^2 ]
                    }
+ \Omega_k(f_t)
\end{equation}
with
\begin{equation}
\Omega_k(f_t) = \gamma T + \frac{1}{2} \lambda \sum_{k=1}^T{ w_k^2 }
\end{equation}
to control the complexity of the tree.
Notice that the number of leaves $T$ and
the scores  of the tree at the leaves $w_k$ ought to be indexed
by $t$ to refer to the $t$-th iteration.

Assuming that the tree structure is known, it is possible to derivative
the optimal weights. We have $f_t(\boldsymbol{x}) = w_{q(\boldsymbol{x})}$
where $q(\boldsymbol{x})$ is the assignment function
which assigns every data point to the $q(\boldsymbol{x})$-th leaf.
Then define $I_j = \{ i | q(\boldsymbol{x}_i) = j \}$,
the indices of all points that end up in the $j$-leaf.
All data points in the same leaf share the same prediction.
Consequently, the objective function can be recast as
\begin{equation}
Obj^{(t)}
=
\sum_{j=1}^T{
                    \left[
                    \left( \sum_{i \in I_j}{ g_i } \right) w_j
                    +
                     \frac{1}{2} \left( \lambda + \sum_{i \in I_j}{ h_i } \right) w_j^2
                   \right]
                   }
+ \gamma T
\end{equation}
We can solve for the optimal weights $w_j^*$: Setting $G_i = \sum_{i \in I_j}{ g_i }$
and $H_i = \sum_{i \in I_j}{ h_i }$, we have
\begin{equation}
w_j^* = - \frac{ G_j }{ H_j + \lambda }
\end{equation}
and so
\begin{equation}
Obj^{(t)}
=
-\frac{1}{2}
\sum_{j=1}^T{
                    \frac{ G_j^2 }{ H_j + \lambda }
                   }
+ \gamma T
\end{equation}

The tree is grown in a top-down recursive fashion:
For each feature in turn, XGBoost sorts the numbers
and scans the best splitting point.
The change of objective after a split is the \emph{gain}:
\begin{equation}
gain
=
\frac{1}{2}
\left[
- \frac{ (G_L+G_R)^2}{H_L+H_R+\lambda}
+
\left(
  \frac{ G_L^2}{H_L+\lambda}
+ \frac{ G_R^2}{H_R+\lambda}
\right)
\right]
- \gamma
\end{equation}
The tree is grown
until the maximum depth is reached.
Then nodes with a negative gain are pruned out in a bottom-up fashion.

\section{Linear models at the leaves}

Instead of a score $w$, a leaf stores the weights of a linear model,
$\boldsymbol{w} \in \mathbb{R}^{d+1}$ where $d$ is the dimension of the
feature space, such that the local prediction
at an unseen input $\boldsymbol{x}_*$ at the leaf is given by
$\boldsymbol{w}^T \tilde{\boldsymbol{x}}_*$
where $\tilde{\boldsymbol{x}}_* = [ \boldsymbol{x}_* \ , \ 1]^T$.

This extension is natural:
\begin{itemize}
\item The simple model $\boldsymbol{w}^T \tilde{\boldsymbol{x}}$ captures
both the linear and constant models.
\item It is a form of \emph{locally weighted polynomial regression}.
However, instead of having a bandwidth controlling how much of the data is
used to fit each local polynomial, all data points inside a specific
hypervolume (the volume of feature space assigned to a leaf) are
considered for model building. As in \emph{mixture of experts},
the input space is divided into regions within which specific separate experts
make predictions.
\end{itemize}

The objective at round $t$  changes from \ref{obj at round t} to
\begin{equation}
Obj^{t)}
=
\sum_{i=1}^n{
  \left[ g_i \tilde{\boldsymbol{x}}_i^T \boldsymbol{w}_{q(\boldsymbol{x}_i)} + \frac{1}{2} \boldsymbol{w}_{q(\boldsymbol{x}_i)}^T h_i \tilde{\boldsymbol{x}}_i \tilde{\boldsymbol{x}}^T_i  \boldsymbol{w}_{q(\boldsymbol{x}_i)} \right]
  }
+ \gamma T + \frac{1}{2} \lambda \sum_{j=1}^T{ \boldsymbol{w}_j^T \boldsymbol{w}_j  }
\end{equation}
Regrouping by leaf, we get
\begin{equation}
Obj^{t)}
=
\sum_{j=1}^T{
  \left[
   \left(
   \sum_{i \in I_j} { g_i \tilde{\boldsymbol{x}}_i^T }
    \right)
   \boldsymbol{w}_j
 +
   \frac{1}{2} \boldsymbol{w}_j^T
     \left(
     \lambda \boldsymbol{I}_{d+1}
     + \sum_{i \in I_j}{  h_i \tilde{\boldsymbol{x}}_i \tilde{\boldsymbol{x}}^T_i }
      \right)
     \boldsymbol{w}_j
  \right] }
+ \gamma T
\end{equation}
By defining
\begin{equation}
\tilde{\boldsymbol{g}}^T_j = \sum_{i \in I_j}{ g_i \tilde{\boldsymbol{x}}_i^T }
\ , \ \
\tilde{\boldsymbol{H}}_j = \sum_{i \in I_j}{  h_i \tilde{\boldsymbol{x}}_i \tilde{\boldsymbol{x}}^T_i  }
\end{equation}
we have:
\begin{equation}
Obj^{t)}
=
\sum_{j=1}^T{
  \left[
   \tilde{\boldsymbol{g}}^T_j \boldsymbol{w}_j +
   \frac{1}{2} \boldsymbol{w}_j^T
               ( \lambda \boldsymbol{I}_{d+1} + \tilde{\boldsymbol{H}}_j )
     \boldsymbol{w}_j
  \right]
+ \gamma T
  }
\end{equation}
Assuming the structure of the tree fixed, the optimal weights are given by
\begin{equation}
\boldsymbol{w}_j^*
= - ( \lambda \boldsymbol{I}_{d+1} + \tilde{\boldsymbol{H}}_j )^{-1} \tilde{\boldsymbol{g}}_j
\end{equation}
We have $\tilde{\boldsymbol{g}}_j = \tilde{\boldsymbol{X}}_j^T \boldsymbol{g}_j$
where $\tilde{\boldsymbol{X}}_j^T$ is the $(d+1)$-x-$|I_j|$ matrix
whose columns are the $\tilde{\boldsymbol{x}}_i$.
From the dyadic expansion, we can see  that
$\tilde{\boldsymbol{H}}_j = \tilde{\boldsymbol{X}}_j^T \boldsymbol{H}_j \tilde{\boldsymbol{X}}_j$
where $\boldsymbol{H}_j$ is a diagonal matrix with elements
the $h_i$, $i \in I_j$.
Because $h_i$ is the second derivative of a loss function, it is positive.
Hence $\tilde{\boldsymbol{H}}_j$ is symmetric positive semi-definite and
$\tilde{\boldsymbol{C}}_j = \lambda \boldsymbol{I} + \tilde{\boldsymbol{H}}_j$
is symmetric positive definite as long as $\lambda > 0$.
If so, $ \tilde{\boldsymbol{C}}_j$ is Cholesky decomposed to solve
for the optimal weights $\boldsymbol{w}_j^*$.
If $\lambda = 0$, then $\tilde{\boldsymbol{C}}_j$ is rank at most $d$,
and so degenerate, if $|I_j|<d+1$
(less than $d+1$ input points fall into the leaf).
In that particular situation, we fall back to the piecewise constant model.

$\tilde{\boldsymbol{C}}_j$ is a $(d+1)$-x-$(d+1)$ matrix
where $d$ is the number of features: $\boldsymbol{x} \in \mathbb{R}^d$.
If $d$ is much larger than the cardinality of $I_j$,
then it might be convenient to use the Woodbury formula. Indeed, we have
\begin{equation}
\tilde{\boldsymbol{H}}_j
=
(\tilde{\boldsymbol{X}}_j^T \boldsymbol{H}_j^{\frac{1}{2}})
(\tilde{\boldsymbol{X}}_j ^T \boldsymbol{H}_j^{\frac{1}{2}})^T
=\tilde{\boldsymbol{X}}_j^h \tilde{\boldsymbol{X}}_j^{hT}
\end{equation}
Using the Woodbury formula, we get:
\begin{equation}
( \lambda \boldsymbol{I}_{d+1} + \tilde{\boldsymbol{H}}_j )^{-1}
=
\frac{1}{\lambda} \boldsymbol{I}_{|I_j|}
-
\tilde{\boldsymbol{X}}^h \left( \boldsymbol{I}_{|I_j|}
+
\frac{1}{\lambda} \tilde{\boldsymbol{X}}^{hT} \tilde{\boldsymbol{X}}^h \right)^{-1} \tilde{\boldsymbol{X}}^{hT}
\end{equation}
Now the matrix to invert has size $|I_j|$-x-$|I_j|$.

For the square loss, the optimal weight is
the regularized least-square solution of the residual at a given leaf.
Indeed, we have
\begin{equation}
\boldsymbol{w}_j^*
= \left( \frac{\lambda}{2} \boldsymbol{I}_{d+1} + \tilde{\boldsymbol{X}}_j^T \tilde{\boldsymbol{X}}_j \right)^{-1} \tilde{\boldsymbol{X}}_j^T ( \boldsymbol{y}_j-\hat{\boldsymbol{y}}_j^{(t-1)})
\end{equation}

The objective to minimize is now
\begin{equation}
Obj^{(t)}
=
-\frac{1}{2}
\sum_{j=1}^T{
   \tilde{\boldsymbol{g}}_j^T
   ( \lambda \boldsymbol{I}_{d+1} + \tilde{\boldsymbol{H}}_j )^{-1}
  \tilde{\boldsymbol{g}}_j }
+ \gamma T
\end{equation}

As in XGBoost, we exhaustively search at each node
for the best axis-aligned split, namely the
split associated with the maximum gain:
\begin{eqnarray}
gain
&=&
\frac{1}{2}
\left[
\tilde{\boldsymbol{g}}_{L,R}^T
\boldsymbol{w}_{L,R}^*
-
\left(
\tilde{\boldsymbol{g}}_{L}^T
\boldsymbol{w}_{L}^*
+
\tilde{\boldsymbol{g}}_{R}^T
\boldsymbol{w}_{R}^*
\right)
\right] \nonumber - \gamma
\end{eqnarray}

This extension is computationally intensive
even though the matrices to invert are small
provided that $d$ is small.
One possibility to drastically reduce the compute at
the expense of accuracy is to use the vanilla XGBoost
at the beginning and switch to linear models at the nodes
once the number of elements in nodes is lower than a user-defined threshold.
Notice though that it is expected that fewer trees will be needed.

In XGBoost, a tree is grown until the maximum depth is reached.
Then nodes with a negative gain are pruned out in a bottom-up fashion.
Why do we accept negative gains?
In the middle of the tree construction,
the gain might be negative, but then the following gains might be significant.
This is reminiscent of the exploitation vs. exploration
in many disciplines, e.g. Reinforcement Learning:
The best long-term strategy may involve short-term sacrifices.
However, all sacrifices are unlikely to be worth it.
Thus, in LinXGBoost, we investigate all subtrees
starting from nodes with a negative gain
in a top-to-bottom fashion
and the subtrees that do not lead to a decrease of the objective
are pruned out.
Eventually, if a tree has a single leaf and it does not lead to
a decrease of the objective,
then the tree is removed and the tree building process is stopped.

As a result, the maximum depth of the trees is set to a very large value.
The tree depth, and so the model complexity,
is limited by the minimum number of samples per leaf and the
minimum number of samples for a split.
For large datasets, the depth could easily exceed the default maximum depth,
so that the default plays nevertheless a role in combating overfitting.

There is another major difference with XGBoost: The bias term
is not regularized, as is usual in e.g. Ridge Regression,
see \citet{friedman2001elements}.
Consequently, the matrix
$\tilde{\boldsymbol{C}}_j = \lambda \boldsymbol{I} + \tilde{\boldsymbol{H}}_j$
is re-written as
$\tilde{\boldsymbol{C}}_j = \boldsymbol{\Lambda} + \tilde{\boldsymbol{H}}_j$
where $\boldsymbol{\Lambda}=\text{diag}(\lambda, \cdots, \lambda,0)$
is a diagonal matrix with $d+1$ elements.

\section{Experiments}

First of all, we checked that LinXGBoost yields the same results as XGBoost.
This was gauged visually for one-dimensional problems,
and by comparing performance metrics in higher dimensions.

Because models at the leaves in XGBoost are linear,
they capture the constant model, and so
we expect LinXGBoost to produce no worse results that XGBoost
except that LinXGBoost is prone to overfit.
In the limit, we could use in LinXGBoost as much trees as in XGBoost
and drastically regularize LinXGBoost but this would be pointless.
In our experiments, we observed that we could in general
get as good results as XGBoost using at most two trees,
and that past five trees, results were either worse or the gain
in accuracy was negligibly small so that the additional compute
by adding more trees was not compensated.
Hence we searched for the best number of LinXGBoost trees
using up to five trees for a random run of an experiment,
and thereafter the number of LinXGBoost trees was fixed
for all runs of the experiment except when mentioned.

In the experiments,
we compare LinXGBoost to the vanilla XGBoost and to Random Forest,
often used as a starting point in ML competitions for regression.

\textbf{Preprocessing}
Data re-scaling is not necessary for the vanilla XGBoost and Random Forest.
Since our extension is based on decision trees, it ought to work straight
out of the-box. But since at the leaves we have a linear model, it is better
to have zero mean features. Indeed, leaving the regularization term by side,
the matrix $\tilde{\boldsymbol{X}}_j^T \tilde{\boldsymbol{X}}_j$
must be inverted ($\boldsymbol{H}_j$ is constant for the square loss).
If the features have zero mean, then
$\tilde{\boldsymbol{X}}_j^T \tilde{\boldsymbol{X}}_j$
is nothing else than $\text{cov}(\tilde{\boldsymbol{X}}_j,\tilde{\boldsymbol{X}}_j)$,
and so, if the covariates $\tilde{\boldsymbol{X}}_j$ are almost
linearly independent, then
$\tilde{\boldsymbol{X}}_j^T \tilde{\boldsymbol{X}}_j$
is approximately diagonal.
If the features do not have zero mean, depending on the problem,
$\tilde{\boldsymbol{X}}_j^T \tilde{\boldsymbol{X}}_j$
might be severely ill-conditioned.

\textbf{Performance metrics}
The performance on all experiments is assessed with the Normalized Mean Square Error (NMSE)
\begin{equation}
\text{NMSE}=\frac{\sum_{i=1}^n{ (y_i - \hat{y}_i)^2}}{\sum_{i=1}^n{ (y_i - \overline{y})^2}}
\end{equation}
The method of guessing the mean of the test points
has a NMSE of approximately 1.

\begin{figure}[!tbp]
  \centering
  \hspace*{\fill}%
  \subfloat[3D view.]{\includegraphics[width=0.25\linewidth]{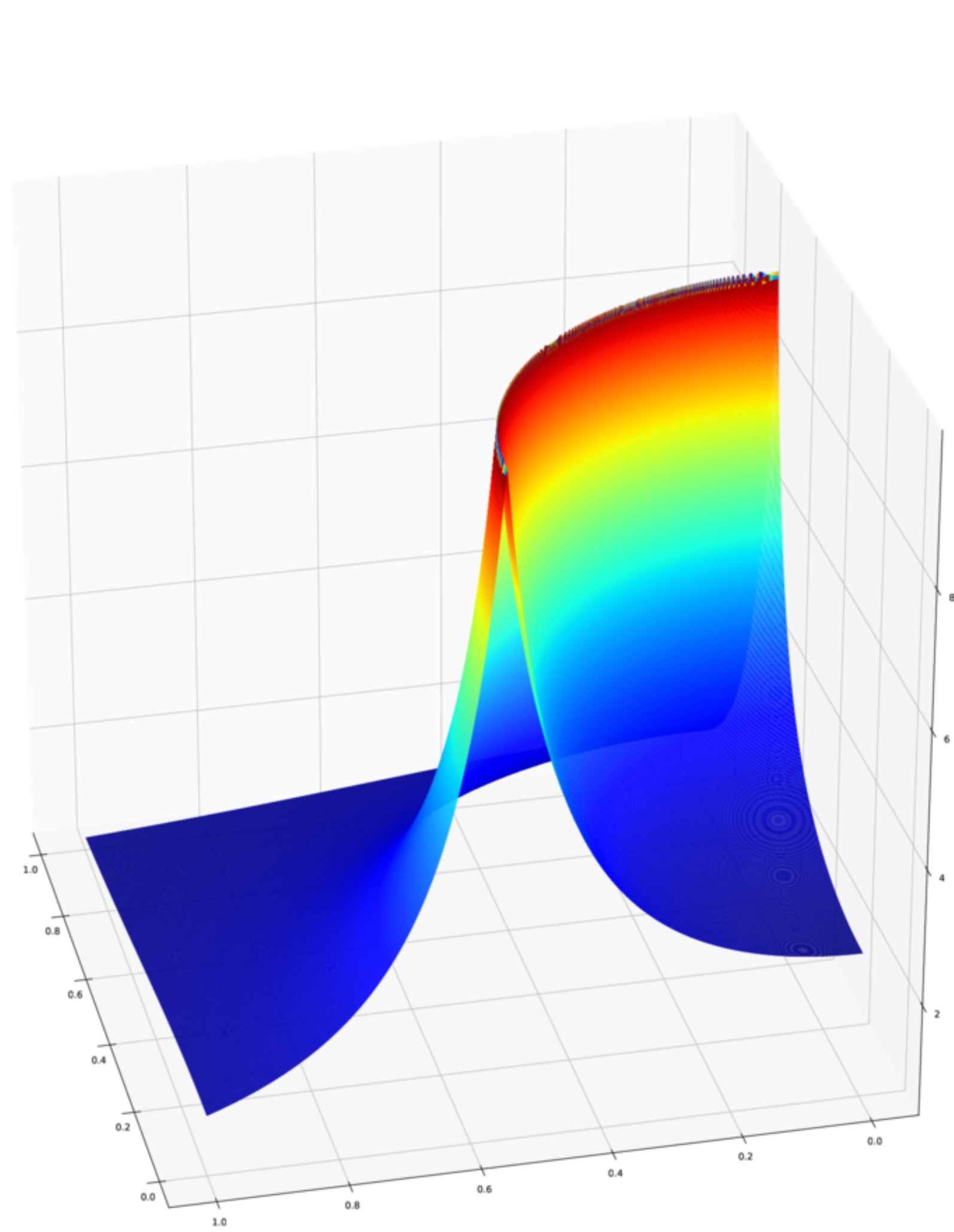}\label{fig:j1 1}}\hfill%
  \subfloat[Projection onto the XY-plane.]{\includegraphics[width=0.25\textwidth]{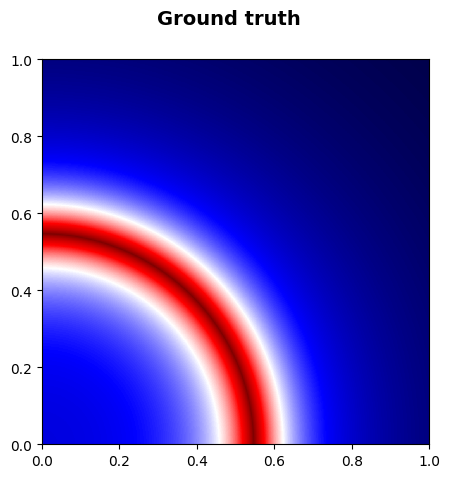}\label{fig:j1 2}}
\hspace*{\fill}%
  \caption{The \emph{Jakeman1} test function.}
  \label{Jakeman1 function}

  \centering
  \subfloat[Training dataset: 11-x-11 noisy samples]{\includegraphics[width=0.48\linewidth]{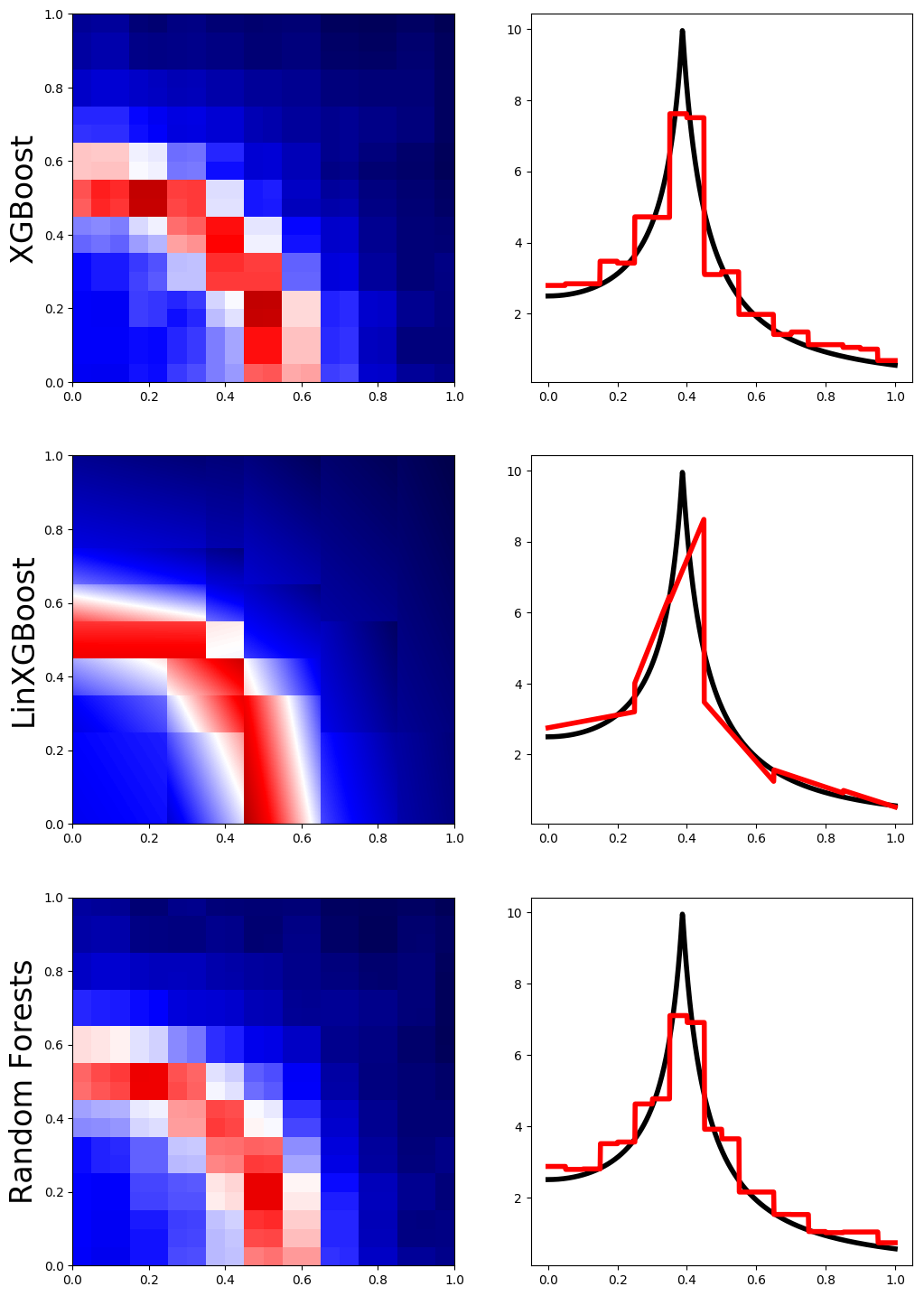}\label{fig:j1 3}}
  \hfill
  \subfloat[Training dataset: 41-x-41 noisy samples]{\includegraphics[width=0.48\linewidth]{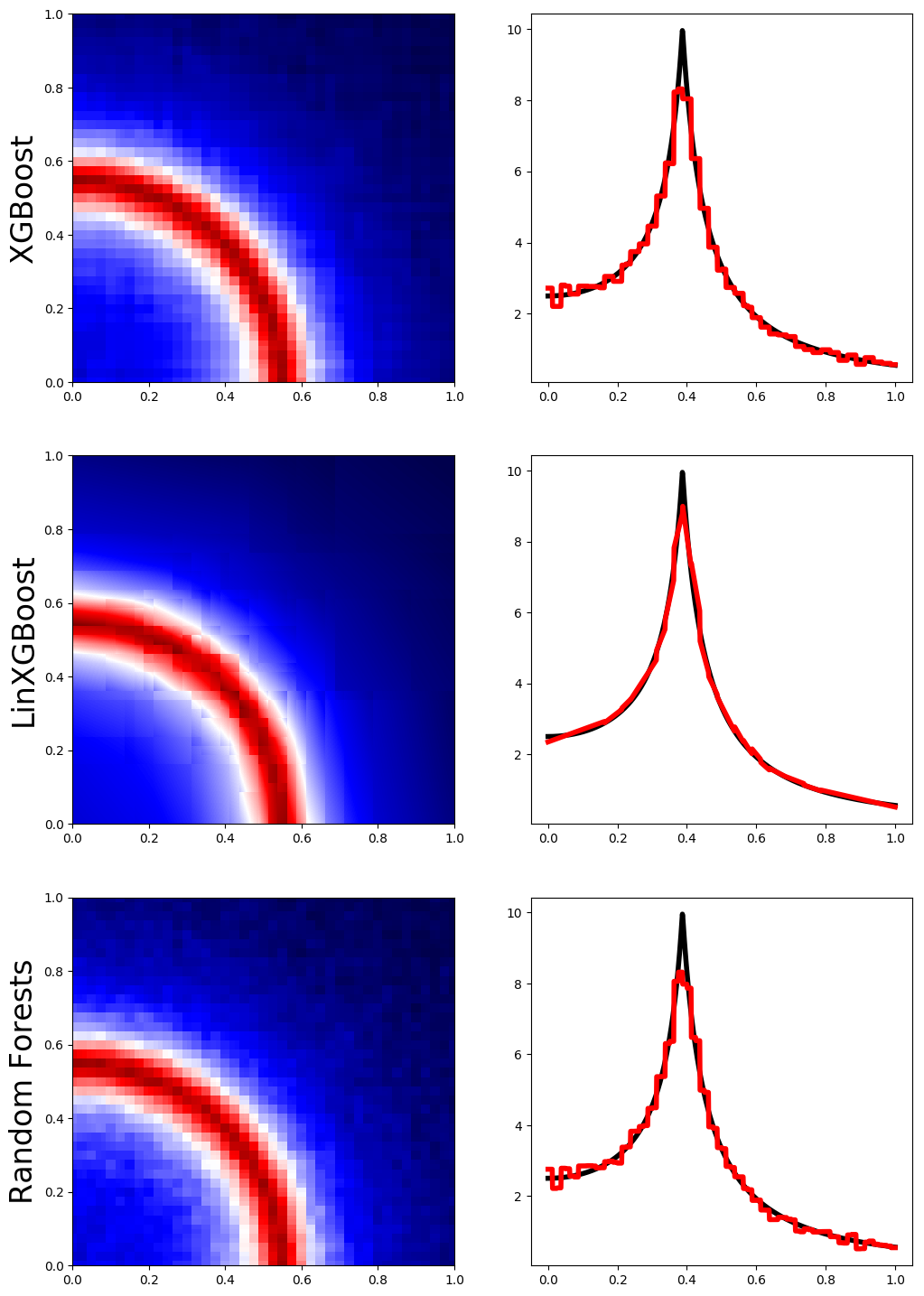}\label{fig:j1 4}}
  \caption{Results for the noisy \emph{Jakeman1} test function ($\sigma^2=0.05$) for a random run
           (see eq. \ref{jakeman1 eq}).
           Training sample data is gridded. Next to the contours, we plot the results
           on the diagonal $x_1=x_2$.}
  \vspace{-20pt}
  \label{Jakeman1 function: results for a random run}
\end{figure}

\begin{figure}[!tbp]
  \centering
  \hspace*{\fill}%
  \subfloat[3D view.]{\includegraphics[width=0.30\linewidth]{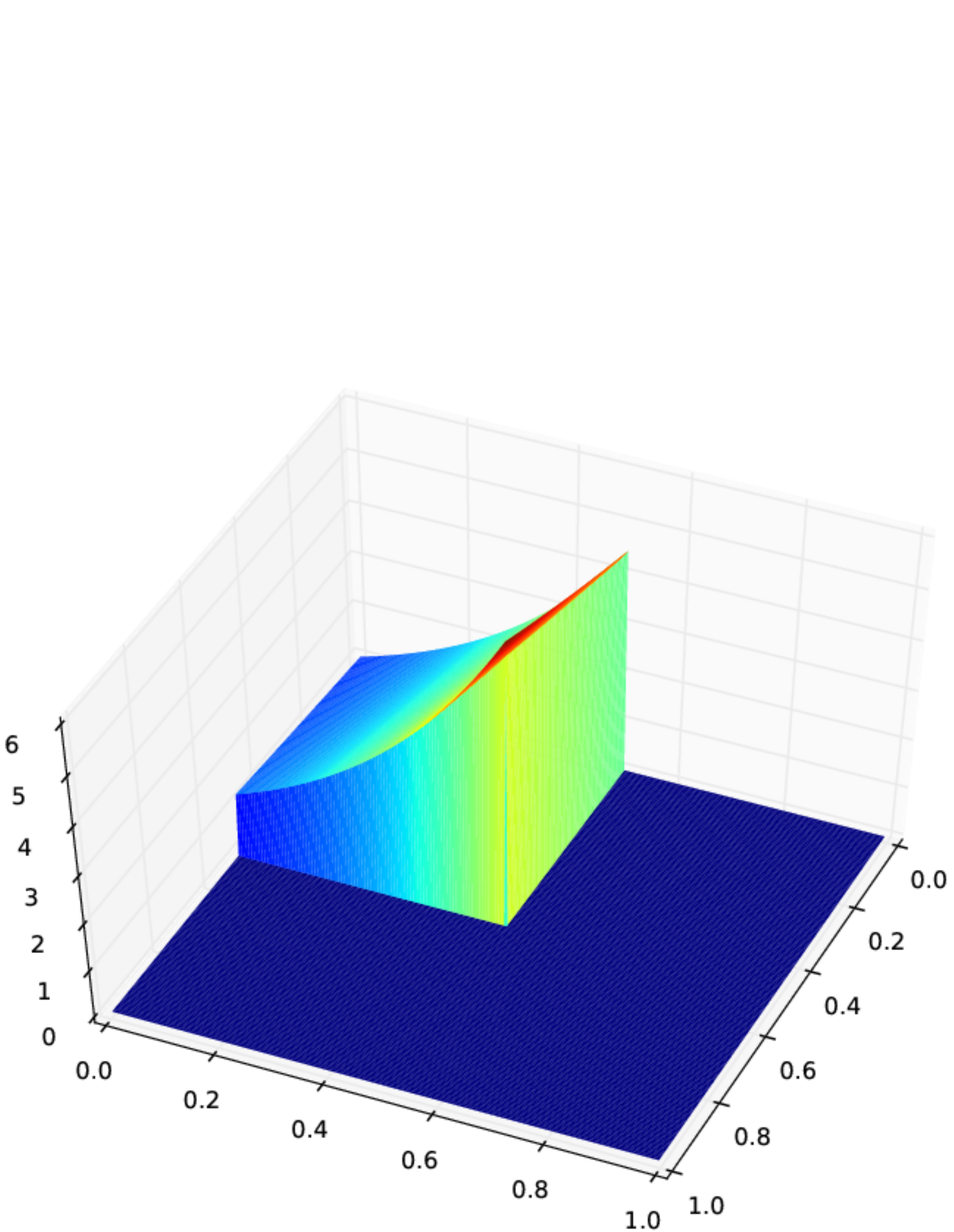}\label{fig:j4 1}}\hfill%
  \subfloat[Projection onto the XY-plane.]{\includegraphics[width=0.25\textwidth]{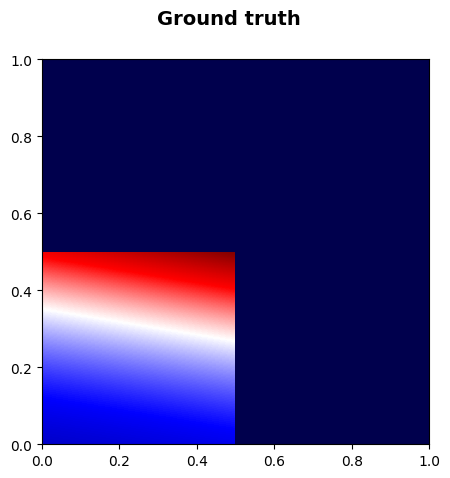}\label{fig:j4 2}}
\hspace*{\fill}%
  \caption{The \emph{Jakeman4} test function.}
  \label{Jakeman4 function}

  \centering
  \subfloat[Training dataset: 11-x-11 noisy samples]{\includegraphics[width=0.48\linewidth]{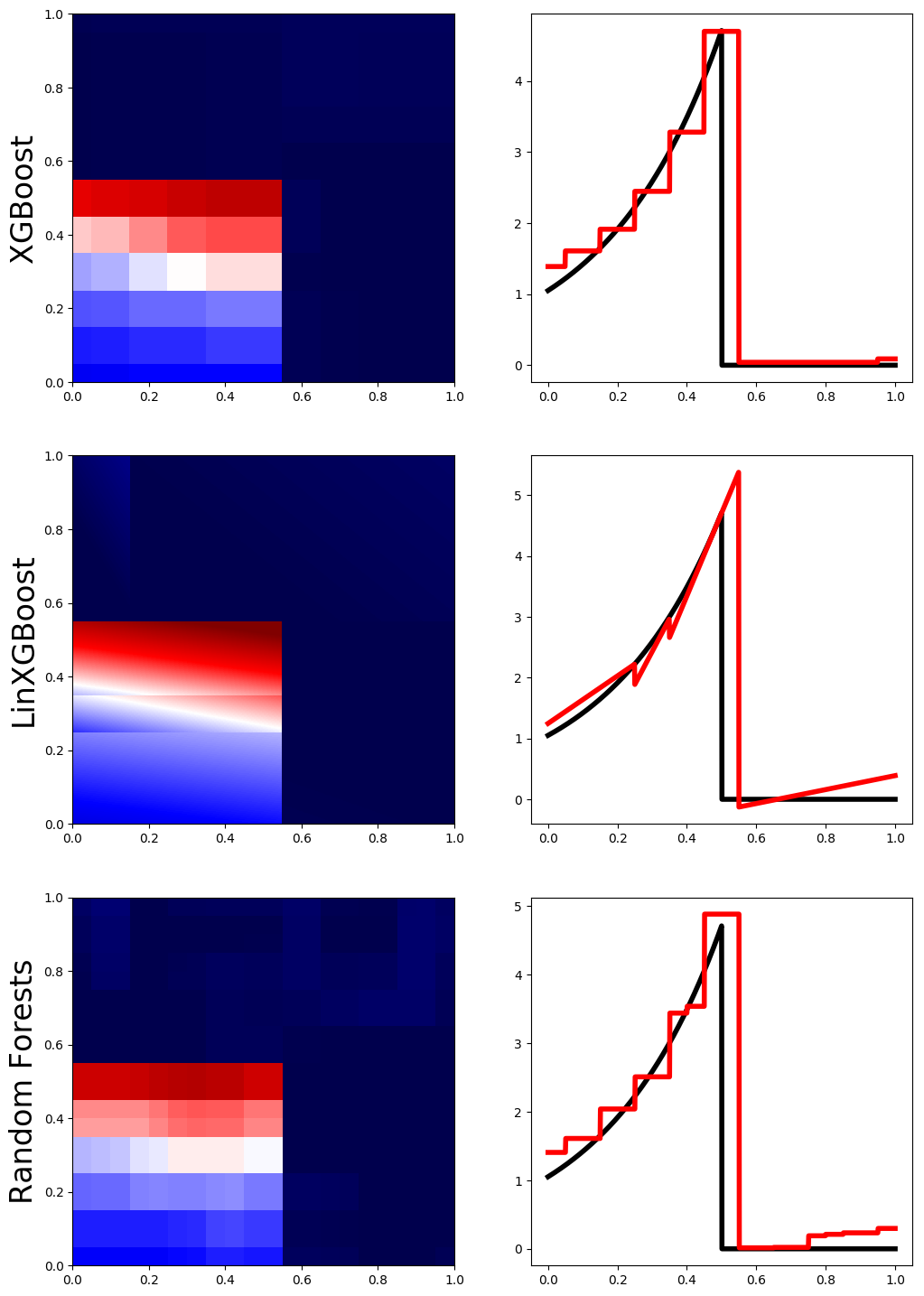}\label{fig:j4 3}}
  \hfill
  \subfloat[Training dataset: 41-x-41 noisy samples]{\includegraphics[width=0.48\linewidth]{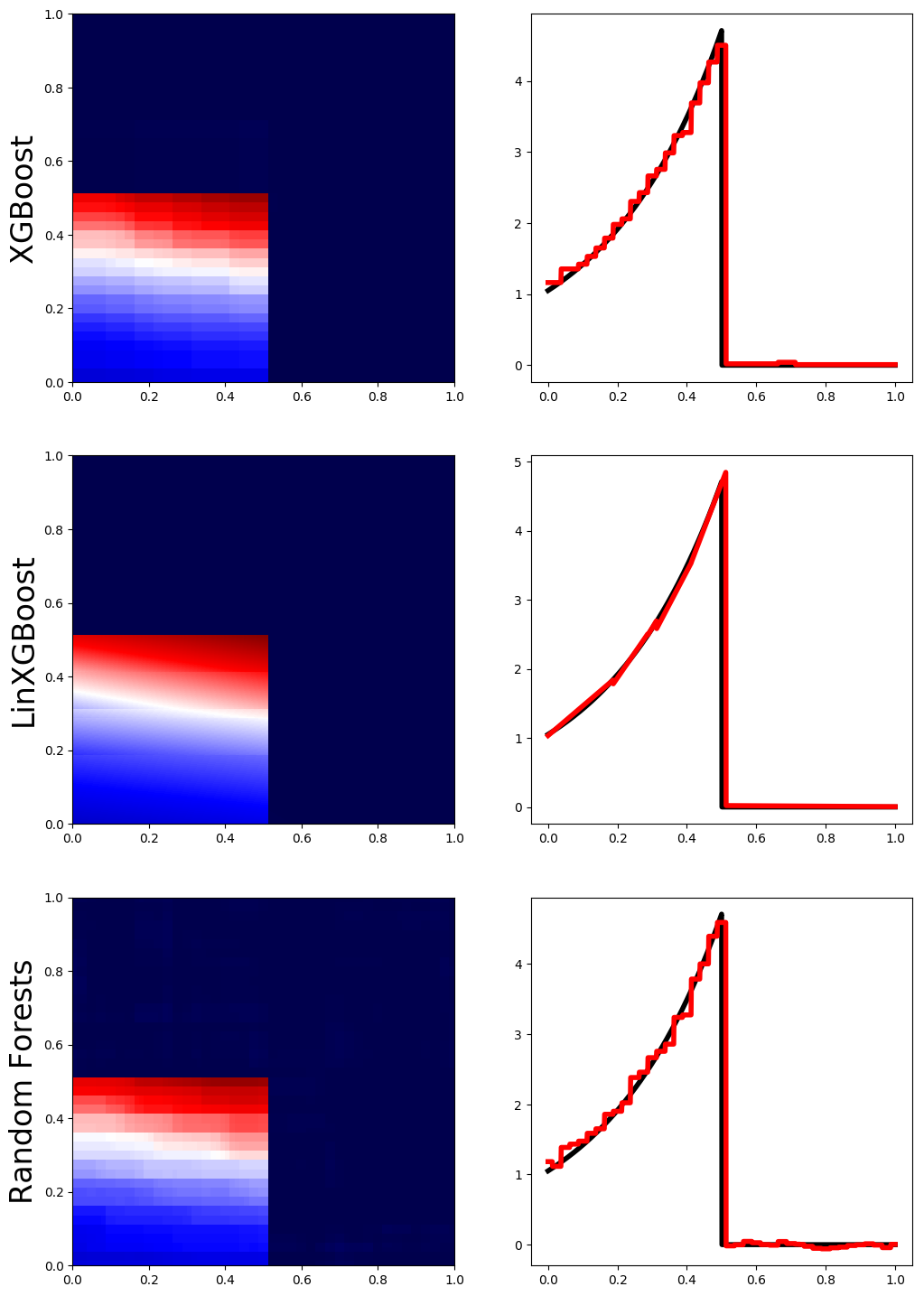}\label{fig:j4 4}}
  \caption{Results for the noisy \emph{Jakeman4} test function ($\sigma^2=0.05$) for a random run
           (see eq. \ref{jakeman4 eq}) .
           Training sample data is gridded. Next to the contours, we plot the results
           on the line $x_1=0.1$.}
  \label{Jakeman4 function: results for a random run}
\end{figure}

\textbf{The Jakeman test functions}
We exercise our model on two synthetic datasets
from \citet{2011arXiv1110.0010J}, both defined on
the unit square. The \emph{Jakeman1} function, Figure \ref{Jakeman1 function},
\begin{equation}
f_1(x_1,x_2)=\frac{1}{|0.3-x_1^2-x_2^2|+0.1}
\label{jakeman1 eq}
\end{equation}
is discontinuous at $x_1^2+x_2^2 = 0.3$, and the \emph{Jakeman4} function,
Figure \ref{Jakeman4 function},
\begin{equation}
f_4(x_1,x_2)
=
\begin{cases} 0 & \text{ if } x_1>0.5 \text{ or } x_2>0.5 \\
\exp(0.5 x_1 + 3 x_2) & \text{otherwise}\end{cases}
\label{jakeman4 eq}
\end{equation}
exhibits a jump at $x_1=0.5, x_2 \in [0,0.5]$ and $x_2=0.5, x_1 \in [0,0.5]$.

To make things interesting, i.i.d. Gaussian noise is added to the training samples.
XGBoost can cope with the noise through subsampling: Only a random fraction
of the training samples pass down a tree. We observed that subsampling plays
a key role when the function exhibits a discontinuity,
otherwise its benefit is minor if the function has a jump.
For subsampling to be effective, a large number of trees must be built.
Hence subsampling degrades the performance of LinXGBoost.
It is better to increase the minimum number of samples at a leaf:
The linear fit to the underlying function is more robust (less susceptible
to be altered by the noise) if we consider more samples.
A positive side-effect is that the trees are shallower and thus
the compute gets faster.

\begin{table}[!ht]
  \caption{NMSE results on the \emph{Jakeman} synthetic test datasets.
  Smaller values are better.}
  \label{sample-table-Jakeman}
  \centering
  \small
  \begin{tabular}{l|c|c|c|c|}
                   & \emph{Jakeman1}       & \emph{Jakeman1}               & \emph{Jakeman4}                 & \emph{Jakeman4}          \\
    Method         & 11-x-11               &  41-x-41                      & 11-x-11                         &  41-x-41                 \\
    \midrule
    XGBoost        &  0.0858 $\pm$ 0.0026  & 0.0083 $\pm$ 0.0002           &  \textbf{0.5778} $\pm$ 0.0214   & 0.1335 $\pm$ 0.0021      \\
    LinXGBoost     &  0.0724 $\pm$ 0.0094  & \textbf{0.0046} $\pm$ 0.0004  &  0.6480 $\pm$ 0.0294            & 0.1385 $\pm$ 0.0024      \\
    Random Forest  &  0.0838 $\pm$ 0.0020  & 0.0113 $\pm$ 0.0002           &  0.5907 $\pm$ 0.0194            & 0.1376 $\pm$ 0.0022      \\
    \bottomrule
  \end{tabular}
  \vspace{-10pt}
\end{table}

Datasets are gridded: Training is carried out on
the small 11-x-11 and medium 41-x-41 configurations,
and testing is done on a 1001-x-1001 configuration.
All model parameters are tuned for each run of an experiment
by conducting an exhaustive grid search
(10-fold cross-validation on the training data set).
An experiment consists of 20 runs.
The number of LinXGBoost trees is fixed, whereas the best number of XGBoost trees
is found by cross-validation. The XGBoost regularizer $\lambda$ was set to 0,
the best setting in all circumstances.
Results are presented in table \ref{sample-table-Jakeman}.

For the \emph{Jakeman1} function, on the small training dataset,
LinXGBoost has 3 trees whereas XGBoost has around 50 trees
(the exact number is run-dependent).
There is no clear-cut winner but we observe that
the variance of LinXGBoost results is fairly high.
On the medium training dataset, LinXGBoost has 5 trees
and XGBoost needs circa 150 trees. Nevertheless, LinXGBoost
beats XGBoost by a large margin. Random Forest performs
slightly worse than XGBoost.
A random run is shown in Figure \ref{Jakeman1 function: results for a random run}.

Again our expectations, LinXGBoost performs slightly worse than XGBoost and Random Forest
on the \emph{Jakeman4} function on the small training dataset.
Results of all methods
are close to each other in terms of NMSE on the medium dataset.
LinXGBoost is limited to 3 trees. Including more trees in LinXGBoost model
does not improve performance.
The main contribution to the error comes from the
jump being at the wrong position: All methods put the jump
far to the right\footnote{The reason therefor is that the training data is gridded.
Indeed, if we repeat the previous experiment with
41x41=1681 input data points randomly distributed in the unit square,
then the error is one order of magnitude lower, and LinXGBoost with 5 estimators
performs as good as XGBoost with 100 estimators.}
The improvement of LinXGBoost in smoother regions is
marginal though visually noticeable.
Both assertions can be verified on the
random run shown in Figure \ref{Jakeman4 function: results for a random run}.

\textbf{The Friedman1 test function}
The \emph{Friedman1} data set is a synthetic dataset.
It has been previously employed in evaluations of MARS,
\citet{friedman1991multivariate},
and bagging, \citet{breiman1996bagging}.
According to Friedman,
it is particularly suited to examine the ability of methods
to uncover interaction effects that are present in the data.
In its basic version, there are 10 independent covariates,
uniformly distributed on (0,1)
and only five of these are related to the target via
\begin{equation}
y=10 \sin(\pi x_1 x_2) + 20 (x_3-0.5)^2 + 10 x_4 + 5 x_5 + \epsilon
\end{equation}
where $\epsilon$ is Gaussian noise with $\sigma=1$.

We consider the dataset from the LIACC repository
that contains 40768 data points.
Results are averaged over 20 randomly splits of this data set.
Each split has 200 training and 40568 testing points.
Model parameters are tuned for each split
by a 10-fold cross-validation on the training data set.

In a random run, we found that LinXGBoost needs 2 to 3 trees
and so for the experiment the number of trees in LinXGBoost
is either 2 or 3 (determined by cross-validation),
whereas XGBoost has approximately 50 estimators.
Results are presented in table \ref{sample-table-Friedman}.
The error of Random Forest is twice as high as the error of XGBoost,
and LinXGBoost does even better than XGBoost on average, but
LinXGBoost results are a bit more volatile,
presumably because LinXGBoost uses very few trees.

\begin{table}[ht]
  \caption{NMSE results on the \emph{Friedman1} synthetic test dataset.
  Smaller values are better.}
  \label{sample-table-Friedman}
  \centering
  \small
  \begin{tabular}{l|c|}
    Method         &  \emph{Friedman1} \\
    \midrule
    XGBoost        &  0.1303 $\pm$  0.0138 \\
    LinXGBoost     &  \textbf{0.1133} $\pm$  0.0199 \\
    Random Forest  &  0.2278 $\pm$  0.0197 \\
    \bottomrule
  \end{tabular}
\end{table}

\textbf{The Combined Cycle Power Plant Data Set (CCPP)}
The dataset, available at the UCI Machine Learning Repository,
contains 9568 data points collected from a Combined Cycle Power Plant
over 6 years (2006-2011),
when the power plant was set to work with full load.
Features consist of
hourly average ambient variables Temperature (T),
Ambient Pressure (AP),
Relative Humidity (RH) and exhaust Vacuum (V)
to predict the net hourly electrical energy output (PE)
of the plant.

Among all datasets used for the benchmark,
this dataset is the only real-world dataset
that was not purposely generated for the validation of algorithms.
As such, we cannot rule out the presence of outliers.
Outliers have a detrimental impact on the results of XGBoost and LinXGBoost
because both make use of the square loss function.
Nevertheless, this impact is attenuated by the fact that
XGBoost and LinXGBoost
are based on decision trees and so outliers are isolated
into small clusters.
Extreme values do not affect the entire model because of local model fitting.

The summary of the data set per variable is:
\begin{verbatim}
                AT            V           AP           RH           PE
count  9568.000000  9568.000000  9568.000000  9568.000000  9568.000000
mean     19.651231    54.305804  1013.259078    73.308978   454.365009
std       7.452473    12.707893     5.938784    14.600269    17.066995
min       1.810000    25.360000   992.890000    25.560000   420.260000
25%      13.510000    41.740000  1009.100000    63.327500   439.750000
50%      20.345000    52.080000  1012.940000    74.975000   451.550000
75%      25.720000    66.540000  1017.260000    84.830000   468.430000
max      37.110000    81.560000  1033.300000   100.160000   495.760000
\end{verbatim}

All features are real-valued (no categorical features).
Furthermore, there are no missing values.

Are there \emph{extreme outliers} ?
Any point beyond 2*step where step=1.5(Q3-Q1)
is considered an extreme outlier, whereas a point beyond
a step is considered a
\emph{mild outlier} \footnote{See
\href{http://www.itl.nist.gov/div898/handbook/prc/section1/prc16.htm}{What are outliers in the data?}
from NIST/SEMATECH e-Handbook of Statistical Methods.}.
Recall that Q1 and Q3 are the 25th and 75th percentile of the data
for a given variable respectively.
Answer: No.

\begin{figure}[h]
  \centering
  \includegraphics[width=15.0cm]{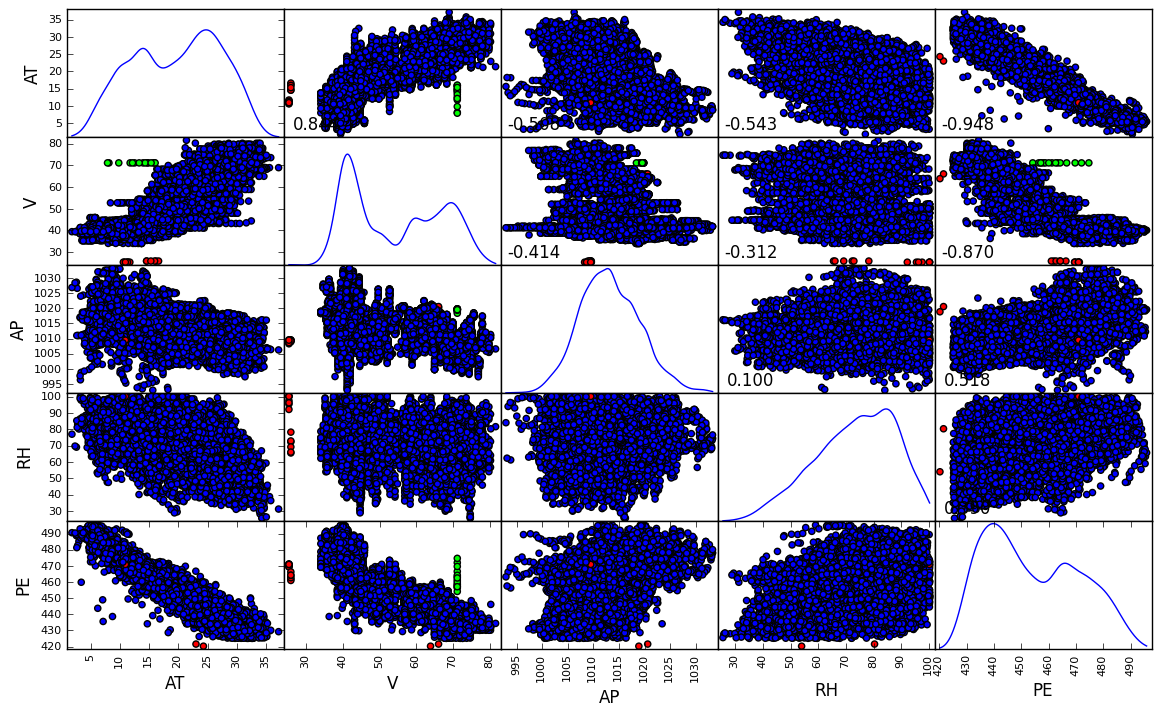}
  \caption{Scatter-plot matrix for the CCPP dataset. Points that are not colored blue are considered outliers.}
  \label{fig: scatter-plot matrix}
\end{figure}

Are there \emph{multivariate outliers} ?
A scatter-plot matrix is an appropriate tool to rapidly
scrutinize many variables for patters (for linear trends,
the correlation coefficients are printed) and outliers.
Figure \ref{fig: scatter-plot matrix} clearly reveals some abnormalities:
\begin{itemize}
\item There are 13 points at the lower end of variable V (colored red) that
are away from the bulk of points (colored blue): 7 points for which V=25.36
and 6 points for which V=25.88.
\item There are 2 isolated points (colored red) at the lower end of variable PE.
\item There is a strip of 15 points (colored green) for which V=71.14,
which indicates that the data is likely to have been thresholded.
\end{itemize}
All those points can all the more safely be removed from the data set
since we have plenty of data points.

Another key property of the data set depicted
in Figure \ref{fig: scatter-plot matrix} is that the target PE seems to be
a (highly noisy) linear function of variable AT: The coefficient of correlation tops
at -0.95. Hence a linear regression is a worthy contender to XGBoost, LinXGBoost
and Random Forest for this particular problem.

The 9568 records set is split into training and test sets in a 70-30 proportion.
The split is repeated 20 times independently. For each split,
a 3-fold cross-validation  is carried out to find the best hyperparameters.

Results are presented in table \ref{real-world-datasets}.
XGBoost wins by a large margin and Random Forest performs slightly better
than LinXGBoost with 4 trees. Results from the linear regression
are fairly good and define the base line,
but it is no surprise since the relationship
between target and features is almost linear.

\textbf{Pumadyn-8nm} The data set was generated using a robot-arm simulation.
It is highly non-linear and has very low noise. It contains 8192 data samples
with 8 attributes. We follow exactly the same procedure as with the
CCPP data set.
Table \ref{real-world-datasets} shows that XGBoost with 150 estimators (set variable)
provides slightly better results than Random Forest with 100 estimators (also set variable)
and LinXGBoost with 3 estimators.

\begin{table}[ht]
  \caption{NMSE results on the CCPP (Combined Cycle Power Plant) and pumadyn-8nm datasets.
  Smaller values are better.}
  \label{real-world-datasets}
  \centering
  \small
  \begin{tabular}{l|c|c|}
    Method         &  CCPP                   & pumadyn-8nm \\
    \midrule
    Linear reg.    &  0.07133 $\pm$  0.00283   & 0.48262 $\pm$ 0.01152              \\
    XGBoost        &  \textbf{0.03371} $\pm$  0.00233   &  \textbf{0.03737} $\pm$  0.00145  \\
    LinXGBoost     &  0.03567 $\pm$  0.00289   &  0.03876 $\pm$  0.00177  \\
    Random Forest  &  0.03988 $\pm$  0.00260   &  0.03898 $\pm$  0.00138  \\
    \bottomrule
  \end{tabular}
\end{table}

\section{Conclusion}

The gradient boosting algorithm XGBoost was extended into LinXGBoost
so that linear models
are stored at the leaves. Computations are still analytically tractable.

This strategy is supposed to rip benefits in terms of accuracy
in low dimensional input space but it is computationally intensive.
As expected, experiments demonstrate that far fewer trees (in general less than
five) are needed
to get the accuracy XGBoost yields with hundreds of trees.
LinXGBoost seems to shine in one-dimensional problems, but
in higher input dimensions, it does not clearly beat XGBoost or Random Forest.

In a future work, we will investigate whether LinXGBoost can offer
substantial improvements in classification.

\subsubsection*{Code}

The LinXGBoost code is written in Python.
It is not an extension of XGBoost.
Why ?
At the first sight, the XGBoost code can appear arcane because of precisely
what makes XGBoost awesome: It is written in C++
and ported to R, Python, Java, Scala, and more,
it runs on a single machine, Hadoop, Spark, Flink and DataFlow,
and it is highly optimized.
All of this bloats the code and makes it harder to pinpoint the core routines
to change (and to apply the changes without breaking any feature of XGBoost!).
We felt it is far more easier to implement
LinXGBoost from scratch.
LinXGBoost is a naive implementation
in Python in less than 350 human-readable lines based on \citet{chen2014introduction}.
Hence the interpretation of the parameters is obvious:
It complies with \citet{chen2014introduction}.

The LinXGBoost implementation does not strive for speed.
The current version runs only on a single machine and is not multithreaded.
Therefore, we do not mention runtime performance.

The software implementation is made available
at \url{https://github.com/ldv1}.
It is envisaged to extend the XGBoost code in a future work.

\medskip

\bibliographystyle{dinat} 
\bibliography{references}

\end{document}